# Efficient Higher-order Convolution for Small Kernels in Deep Learning

Zuocheng Wen     Lingzhong Guo

University of Sheffield, UK

zwen12@sheffield.ac.uk,  l.guo@sheffield.ac.uk

## Abstract

Deep convolutional neural networks (DCNNs) are a class of artificial neural networks, primarily for computer vision tasks such as segmentation and classification. Many nonlinear operations, such as activation functions and pooling strategies, are used in DCNNs to enhance their ability to process different signals with different tasks. Conceptional convolution, a linear filter, is the essential component of DCNNs while nonlinear convolution is generally implemented as higher-order Volterra filters, However, for Volterra filtering, significant memory and computational costs pose a primary limitation for its widespread application in DCNN applications. In this study, we propose a novel method to perform higher-order Volterra filtering with lower memory and computation cost in forward and backward pass in DCNN training. The proposed method demonstrates computational advantages compared with conventional Volterra filter implementation. Furthermore, based on the proposed method, a new attention module called Higher-order Local Attention Block (HLA) is proposed and tested on CIFAR-100 dataset, which shows competitive improvement for classification task. Source code is available at: https://github.com/WinterWen666/Efficient-High-Order-Volterra-Convolution.git

## 1. Introduction

DCNNs are a class of artificial neural networks that achieve high accuracy in computer vision (CV) field such as image classification, semantic segmentation, and object detection [1-4]. Especially, the shortcut connections in Residual Networks (ResNets) [5-6] solve the vanishing gradient problem in much deeper network architecture that makes deeper networks obtain significantly improved performance. However, conventional convolution used in DCNN is a linear operation that has led to the proposals of various non-linear components (e.g. nonlinear activation functions and pooling strategies etc) [7-8] to introduce the nonlinearity into the networks. It is well known that convolutional neural networks have largely drawn inspiration from natural visual systems. Recent research results of neuroscience have shown the existence of non-linear operations in the response of complex visual cells [9-12], there is a need to incorporate these nonlinear functions/mechanisms into deep learning architecture to reflect our complicated visual systems. Therefore, it is worth some more exploitation in this direction, where higher-order convolution is one of the ways to explore the nonlinearity in deep learning architectures.

Some research works have used second-order terms to improve the generalization ability of the deep learning neural networks [13-15]. Moreover, based on second-order information, some post-processing methods such as Matrix Power Normalized Covariance, Matrix Square Root Normalization, and Second Order Covariance pooling also achieve competitive performance [16-18]. Besides, attention mechanism has brought impressive improvements in deep learning in CV. In general, according to dimensions of inputs, attention methods target on the channel, spatial and branch dimensions by using either sigmoid or SoftMax functions to obtain attention coefficients [19-22]. Furthermore, considering the interactions among spatial, channel, or branch dimensions has resulted in the proposal of methods that consider the interaction of the original inputs, deriving coefficients of self-correlation known as self-attention [23-26]. Essentially, self-attention is one of methods to use the second order information in deep learning models. Therefore, effectively utilizing second order information or higher order information remains a worthwhile problem to investigate. These studies have evidenced the effectiveness of introducing such second-order information (either directly or through some post-processing methods) in the networks in terms of the accuracy whilst the high memory requirement and computation cost hinder their wider use in practice (especially for orders higher than 2). In this study, we present a new algorithm to perform higher-order Volterra-based convolution with lower memory and computation cost, base on which a novel Higher order Local Attention block (HLA) is proposed to enhance the performance of deep learning neural networks for image classification.

## 2. Revisiting conventional implementation of higher order Volterra filters

In this section, we elaborate the forward pass and backward pass of conventional implementation of Volterra filters in the context of backpropagation training. Based on that, we will optimise the backward pass by transposing the weights instead of computing Jacobian matrix.

### 2.1 Forward pass

For a discrete $r$th order Volterra filter [35], suppose the input is a vector $x \in R^n$

$$x = [x_1\ x_2\ \cdots\ x_n]^T \quad (1)$$

and the output is $y^r(x) \in R$, it has the following general expression between the input and output as a filter

$$y^r(x) = \sum_{i_1=1}^{n} w_{i_1} x_{i_1} + \sum_{i_1=1}^{n}\sum_{i_2=1}^{n} w_{i_1,i_2} x_{i_1} x_{i_2} + \cdots$$
$$+ \sum_{i_1=1}^{n}\cdots\sum_{i_r=1}^{n} w_{i_1,i_2\cdots,i_r} x_{i_1} x_{i_2} \cdots x_{i_r} + b \quad (2)$$

where $w_{i_1}$, $w_{i_1,i_2}$, …, $w_{i_1,i_2,\cdots,i_r}$ are the weights of the filter's first, second, …, and $r$th-order terms, and $b$ is the bias. To implement the nonlinear filter efficiently in practice, Eq. (1) is transformed into the following form:

$$y^r(x) = w_{R_1}^T x_{R_1} + w_{R_2}^T x_{R_2} + \cdots + w_{R_j}^T x_{R_j} + b \quad (3)$$

where $w_{R_j}$ and $x_{R_j}$, $j = 1,2,\ldots,r$ are $j$th order weight vector and $j$th order term vector. For $j > 1$, the $j$th order term vector $x_{R_j}$ is defined recursively as follows:

$$x_{R_1} \coloneqq x, x_{R_j} \coloneqq x_{R_{j-1}} \otimes x \quad (4)$$

where $\otimes$ is the Kronecker product. Following the Kronecker product, the $j$th order weight vector is given by

$$w_{R_j} = \left[\underbrace{w_{1,1,\cdots,1}}_{j}\ \underbrace{w_{1,1,\cdots,2}}_{j}\ \cdots\ \underbrace{w_{j,j,\cdots,j-1}}_{j}\ \underbrace{w_{j,j,\cdots,j}}_{j}\right]^T \quad (5)$$

Then, the forward pass of traditional high order Volterra-based convolution can be easily achieved by the process mentioned above.

### 2.2 Backward pass

Consider the learning problem of the neural networks, we want to minimize an error function $E$, which depends on the weight of the Volterra filter. To train Volterra weights and propagate the error, we must compute the gradients of the output $y^r(x)$, with respect to the weights $w_{i_1,i_2,\cdots,i_j}$ and $x$. The gradient with respect to $w_{i_1,i_2,\cdots,i_j}$ is simply $\frac{\partial E}{\partial y^r(x)} \frac{\partial y^r(x)}{\partial w_{i_1,i_2\cdots,i_j}} = \frac{\partial E}{\partial y^r(x)} x_{i_1} x_{i_2} \cdots x_{i_j}$, which is also $\frac{\partial E}{\partial y^r(x)} \frac{\partial y^r(x)}{\partial w_{R_j}} = \frac{\partial E}{\partial y^r(x)} x_{R_j}$. The gradient respects to $x$ is $\frac{\partial E}{\partial y^r(x)} \frac{\partial y^r(x)}{\partial x} = \frac{\partial E}{\partial y^r(x)} (\sum_{j=1}^{r} w_{R_j}^T \frac{\partial x_{R_j}}{\partial x})$, where $\frac{\partial x_{R_j}}{\partial x}$ is the Jacobian matrix of size of $n^j \times n$. However, the computation process by Jacobian matrix will have high time complexity $\mathcal{O}(n^{j+1})$ and auxiliary space $\mathcal{O}(n^{j+1})$. To optimize the computation process, let $W^{(j)}[i_1, i_2 \cdots, i_j] = w_{i_1,i_2\cdots,i_j}, j = 1,2,3,\ldots,r$ be the $j$th order weight matrix of the Volterra filter (2). Note that the general $j$th term of the Volterra convolution is $x_{i_1} x_{i_2} \cdots x_{i_j}$ and its weight is $w_{i_1,i_2,\cdots,i_j}, 1 \leq i_1, i_2 \cdots, i_j \leq n$. Firstly, for $j = 1$, the gradient with respect to $x$ will be

$$\frac{\partial y^r(x)}{\partial x} = W^{(1)} = w_{R_1} = a_1 \quad (6)$$

For $j = 2$, the gradient with respect to $x$ will be

$$\frac{\partial y^r(x)}{\partial x} = a_1 + \left(W^{(2)}[i_1, i_2] + W^{(2)}[i_2, i_1]\right)x_{R_1}$$
$$= a_1 + (a_2[1] + a_2[2])x_{R_1} \quad (7)$$

where $a_2[1] = W^{(2)}[i_1, i_2], a_2[2] = W^{(2)}[i_2, i_1]$. In general, for $j > 2$,

$$a_j \coloneqq \sum_{k=1}^{j} a_j[k] =$$
$$\underbrace{W^{(j)}[i_1,\cdots] + W^{(j)}[i_2, i_1,\cdots] + \cdots + W^{(j)}[i_k, i_1,\cdots]}_{j} \quad (8)$$

where $k = 1,2,\ldots,j$, $W^{(k)}[i_k, i_1,\cdots]$ means move $k_{th}$ dimension to the first dimension of matrix that can be achieved by transpose function in implementation. Please note that if $j \geq 3$, the $a_j$ is a muti-dimension matrix that needs to be transformed into size of $n \times n^{j-1}$ by reshape function in implementation. In general, for $j > 2$, the gradient with respect to $x$ will be

$$\frac{\partial y^r(x)}{\partial x} = a_1 + \sum_{j=2}^{j} a_j x_{R_{j-1}} \quad (9)$$

After the optimization, the time complexity and auxiliary space of the computation process becomes to $\mathcal{O}(n^j)$ and $\mathcal{O}(2n^j)$, respectively.

# 3. Implementing higher-order Volterra filters through positions of input elements

As we discussed in section 2, the complexity of traditional high order convolution calculation will increase exponentially with the order and the number of variables increasing. Although second order convolution can be achieved by upper triangular weighted matrix to reduce the number of weights to $\frac{n(n+1)}{2}$ [14], so far there is no efficient method to perform higher order ($r > 2$) Volterra filtering in deep learning community. Based on our observations, the repeated higher order terms (such as $x_1 x_2$ and $x_2 x_1$) take up massive time and space in the computation process. Therefore, we propose a method that create unique higher order terms to reduce the time and space complexity. In this section, we consider symmetrical kernels as in the common practice with Volterra filter, it follows that we only need to consider the terms of $x_{i_1} x_{i_2} \cdots x_{i_r}$ in (2) satisfying the condition

$$i_1 \leq i_2 \leq \cdots \leq i_r \qquad (10)$$

which, in general form, (2) becomes

$$y^r(\boldsymbol{x}) = \sum_{i_1=1}^{n} w_{i_1} x_{i_1} + \sum_{i_1=1}^{n} \sum_{i_2=i_1}^{n} w_{i_1,i_2} x_{i_1} x_{i_2} + \cdots + \sum_{i_1=1}^{n} \cdots \sum_{i_r=i_{r-1}}^{n} w_{i_1,i_2\cdots,i_r} x_{i_1} x_{i_2} \cdots x_{i_r} + b \qquad (11)$$

However, especially for parallel computation, the implementation of (11) is a problem. To tackle this problem, we propose some algorithms by the position of input to implement equation (11) as followed.

## 3.1 Full Position Matrix (FPM) algorithm with its explanation

To create unique higher order terms of the Volterra filter, we will use positions of the elements of the input to describe these higher order terms. For the input $\boldsymbol{x} = [x_1\ x_2\ \cdots\ x_n]^T \in R^n$, the positions of elements $x_1, x_2, \cdots, x_n$ are simply 1, 2, $\cdots$, $n$. By using the positions, all the terms can be formed. For instance, the term $x_1 x_1 x_3$ can be obtained by accessing to the positions [1, 1, 3] of the input elements $x_1$ and $x_3$ and multiplying them together. In general, we want to obtain all possible position combinations $[i_1 \leq i_2, i_1 \leq i_2 \leq i_3, \ldots, i_1 \leq i_2 \leq \cdots \leq i_r]$. Firstly, we create the no-identical combinations $[i_1 < i_2, i_1 < i_2 < i_3, \ldots, i_1 < i_2 < \cdots < i_r]$. Then, based on no-identical combinations, we create partial-identical and identical combinations to obtain all possible position combinations. For $1^{st}$ order terms $x_1, x_2, \cdots, x_n$ we introduce its No-identical Position Matrix (NPM) as

$$NPM^1 = [\ 1\quad 2\quad 3\quad \ldots\quad n\ ]^T \qquad (12)$$

To create the $NPM^2$ for the second order terms $i_1 < i_2$, we can form a series of vectors by shifting the elements in $NPM^1$ to have

$$\begin{bmatrix} 1 & \ldots & n & 1 & \ldots & n & \ldots & 1 & \ldots & n \\ 2 & \ldots & n+1 & 3 & \ldots & n+2 & \ldots & n & \ldots & 2n \end{bmatrix}^T \qquad (13)$$

Then we obtain the $NPM^2$ by dropping all the rows that contains indices greater than $n$.

$$\begin{bmatrix} 1 & 2 & \ldots & n-1 & 1 & 2 & \ldots & n-2 & \ldots & 1 \\ 2 & 3 & \ldots & n & 3 & 4 & \ldots & n & \ldots & n \end{bmatrix}^T \qquad (14)$$

The whole process is achieved by step 1 to step 14 of Algorithm 1. Consider partial-identical and identical combinations, identical combinations $i_1 = i_2 = \cdots = i_r$ is simply repeating $NPM^1$ $r$ times and partial-identical combinations must define the repeating times of $i_1, i_2, \cdots, i_r$ manually. For example, third order combination has equally position combination $i_1 = i_2 < i_r$ and $i_1 < i_2 = i_r$. To achieve this, we can repeat $i_1$ and $i_2$ of $NPM^2$ twice respectively. For instance, the identical combinations and partial-identical combinations of third order are

$$NPM^3_{i_1=i_2=i_r} = \begin{bmatrix} 1 & 2 & 3 & \ldots & n \\ 1 & 2 & 3 & \ldots & n \\ 1 & 2 & 3 & \ldots & n \end{bmatrix}^T \qquad (15)$$

$$NPM^3_{i_1=i_2<i_r} = \begin{bmatrix} 1 & 2 & \ldots & n-1 & 1 & 2 & \ldots & n-2 & \ldots & 1 \\ 1 & 2 & \ldots & n-1 & 1 & 2 & \ldots & n-2 & \ldots & 1 \\ 2 & 3 & \ldots & n & 3 & 4 & \ldots & n & \ldots & n \end{bmatrix}^T \qquad (16)$$

$$NPM^3_{i_1<i_2=i_r} = \begin{bmatrix} 1 & 2 & \ldots & n-1 & 1 & 2 & \ldots & n-2 & \ldots & 1 \\ 2 & 3 & \ldots & n & 3 & 4 & \ldots & n & \ldots & n \\ 2 & 3 & \ldots & n & 3 & 4 & \ldots & n & \ldots & n \end{bmatrix}^T \qquad (17)$$

$$PM^3_{full} = concat\{NPM^3_{i_1=i_2=i_r}, NPM^3_{i_1=i_2<i_r}, NPM^3_{i_1<i_2=i_r}\} \qquad (18)$$

where $concat\{\ldots\}$ means concatenate the matrices and $PM^3_{full}$ is full position matrix of third order. To integrate this process, we introduce Repeating Matrix (RM) which is all possible repeating position with its repeating times. The creation process of RM can be regarded as Integer Ordered splitting problem where Integer is the highest order of Volterra series. The shapes of RM are designed as 1×1, 1×2, 1×3, …, 1× Integer. Then, splitting the Integer followed the shape of RM and placing them into corresponding RM with ordered sequence. For instance, the third order RMs are [3], [1, 2], [2, 1] and [1, 1, 1]. Then concatenate RM together to create Total Repeating Matrices (TRM). For instance, $TRM^3$ =[[3], [1,2], [2,1], [1, 1, 1]] (this is matrix concatenation, it is not [3, 1, 2, 2, 1, 1, 1, 1]) and $TRM^4$ =[[4], [1,3], [3,1], [2, 2], [2, 1, 1], [1, 2, 1], [1, 1, 2], [1, 1, 1, 1]]. The size of each 1D matrix in $TRM^r$ corresponds to the $r$th $NPM$. For example, $TRM^3$ corresponds to $[NPM^1, NPM^2, NPM^2, NPM^3]$. Therefore, to obtain all possible position of elements used in higher order convolution, we create a list $TNPM^r = [NPM^1, NPM^2, ..., NPM^r]$ for corresponding to $TRM^r$. The process mentioned above is achieved by step 15 to step 18 of Algorithm 1. Based on the idea, we create Full Position Matrix (FPM) to represent all unique higher order terms. The creation of FPMs only needs to meet the condition (10). We present the following algorithm to create FPMs.

---

**Algorithm 1: Creation of Full Position Matrix (FPM)**

**Inputs**: Order: $r$, Kernal-size: $[k_1, k_2]$, Total Repeating Matrices: $TRM^r$, Total No-identical Position Matrix $TNPM^r \leftarrow [\ ]$

**Output**: Full Position Matrix: $PM_{full}^r \leftarrow [\ ]$ for initiation

1. $NPM^r \leftarrow [0, 1, 2, ..., k_1 k_2]^T$ // Position matrix of first order
2. $TNPM^r \leftarrow$ append $(NPM^r)$
3. **for** $i \leftarrow 0$ **to** $r - 1$ **do**
4.    $N_c \leftarrow NPM^r.shape[1]$ // Get the number of columns of $NPM^r$
5.    **for** $j \leftarrow 0$ **to** $k_1 k_2 - N_c$ **do**
6.      $j \leftarrow j + 1$
7.      $NP^r \leftarrow NPM^r[:, N_c - 1] + j$ // Position of next order elements to multiply
8.      $Chosen \leftarrow argwhere(NP^r < k_1 k_2)$ // Find the index of valid position
9.      PM-Ind $\leftarrow Concatenate(NPM^r[Chosen[:,0]], NP^r[Chosen[:,0]])$
     // Obtain the position matrix of high order terms
10.      $NPM \leftarrow$ append (PM-Ind)
11.      $NP^r \leftarrow NP^r - j$ // Recover the $NP^r$
12.    $NPM^r \leftarrow$ vertical stack$(NPM)$
13.    $TNPM^r \leftarrow$ append $(NPM^r)$
14.    $NPM \leftarrow [\ ]$
15. **for** $k \leftarrow 0$ **to** $len(TRM^r)$ **do**
16.    $PM \leftarrow TNPM^r[len(TRM^r[K]) - 1)]$ // Accesses to $k_{th}$ order no-identical PM
17.    $PM_{full}^r \leftarrow$ append $(repeat\ (PM, TRM^r[K], axis = 1))$
// Append identical and partial-identical combinations
18. $PM_{full}^r \leftarrow$ vertical stack$(PM_{full}^r)$

**Return** $PM_{full}^r$

---

The generic type to compute the total number of parameters, $nV$, for a Volterra filter of order $r$ is:

$$nV = \frac{(n+r)!}{n!\,r!} = \binom{n+r}{r} \qquad (19)$$

In $PM_{full}^r$, the number of rows equals to the number of higher order terms ($\frac{(n+r-1)!}{r!(n-1)!}$). The total number of parameters and higher order terms is

$$nV = \sum_{r=1}^{r} \frac{(n+r-1)!}{r!\,(n-1)!} + 1 = \frac{(n+r)!}{n!\,r!} \qquad (20)$$

where 1 is the number of bias.

To prove this equation is correct by induction, $r+1$ must also satisfy the equation $\sum_{r=1}^{r=r+1} \frac{(n+r-1)!}{r!(n-1)!} + 1 = \frac{(n+r+1)!}{n!(r+1)!}$. It follows that

$$\sum_{r=1}^{r=r+1} \frac{(n+r-1)!}{r!\,(n-1)!} + 1 = \sum_{r=1}^{r=r} \frac{(n+r-1)!}{r!\,(n-1)!} + 1$$

$$+ \frac{(n+r)!}{(r+1)!\,(n-1)!} = \frac{(n+r)!}{n!\,r!} + \frac{(n+r)!}{(r+1)!\,(n-1)!}$$

$$= \frac{(r+1)(n+r)! + n(n+r)!}{n!\,(r+1)!}$$

$$= \frac{(n+r+1)!}{n!\,(r+1)!} \qquad (21)$$

### 3.2 High-order convolution by FPMs

Based on Algorithm 1, the simplest method to obtain higher order terms is to access to all the corresponding elements of the input $x$ according to each column of $PM_{full}^r$ and multiplying them to form the terms $x_{S_r}$. However, the process has a high time complexity when the order is higher than 2, because the process is not progressive computation from lower order terms to higher order terms. Here, we introduce an optimized implementation to accomplish progressive computation. Before demonstrating our optimized implementation, we clarify some properties of $PM_{full}^r$.

- Property 1: $x_{S_r}$ and $PM_{full}^r$ is one-to-one correspondence.
- Property 2: In $PM_{full}^r$, rows of any combination of $r - 1$ columns only contain rows of $PM_{full}^{r-1}$.
- Property 3: Swap the specified columns of $PM_{full}^r$ means transpose specified the dimensions of corresponding muti-dimensional sparse matrix.

According to Property 1, the index of rows in $PM_{full}^{r-1}$ represents combinations position of $PM_{full}^{r-1}$, which also represents position of $x_{S_{r-1}}$. Based on property 2, any combination of $r-1$ columns of $PM_{full}^r$ is used for finding all indices of rows in $PM_{full}^{r-1}$ existing in $PM_{full}^r$. Then, combine index of rows and one column that is not component of a combination as $r_{th}$ Progressive Computation Matrix ($PCM^r$). Finally, $x_{S_r}$ can be obtained progressively by

$$x_{S_r} = x[PCM^r[:,1]] \odot x_{S_{r-1}}[PCM^r[:,2]] \quad (22)$$

where $\odot$ is Hadamard product, $PCM^r[:,1]$ is one column that is not component of the combination and $PCM^r[:,2]$ is the indices of rows in $PM_{full}^{r-1}$ existing in $PM_{full}^r$. $x[...]$ means accessing to the elements of $x$. Then, the forward pass can be calculated similar to (3)

$$y^r(x) = w_{S_1}^T x_{S_1} + w_{S_2}^T x_{S_2} + \cdots + w_{S_r}^T x_{S_r} + b \quad (23)$$

As we stated above, the creation of PCMs is

---

**Algorithm 2: Creation of PCMs**

**Inputs**: $PM_{full}^{r-1}$, $PM_{full}^r$, $PCM^r \leftarrow [\,]$ for initiation
**Outputs**: $PCMs^r$: $PCMs^r \leftarrow [\,]$ for initiation
**for** $i \leftarrow 0$ **to** $PM_{full}^r.shape[1]$ **do**
  $PM_{delete} \leftarrow Concat(PM_{full}^r[:,:i], PM_{full}^r[:,i+1:])$ // Delete $PM_{full}^r[:,i]$
  **for** $row\_r$ **in** $PM_{delete}^r$ **do**
    // Find the position of high order terms in $PM_{full}^{r-1}$
    $Position \leftarrow where\,(PM_{full}^{r-1} = row\_r, axis = 1)$
    $PCM^r \leftarrow append(Position)$
  $PCM^r \leftarrow Vstack(PCM^r)$
  $PCM^r \leftarrow Horizontal\ concatnate\,(PM_{full}^r[:,i], PCM^r)$
  $PCMs^r \leftarrow append(PCM^r)$
  $PCM^r \leftarrow [\,]$
**Return** $PCMs^r$

*Please note this algorithm will be used when $r \geq 3$. The $PCMs^2$ is $PM_{full}^2$ and swapped column of $PM_{full}^2$.*

---

The output $PCMs^r$ of the algorithm have $r$ numbers of $PCM^r$ and any $PCM^r$ of $PCMs^r$ can be used in forward pass. The reason we create $r$ numbers of $PCM^r$ is that those $PCM^r$ is used in backward pass. As we mentioned in section 2, the backward pass can be accomplished by summing transposed weights matrix. Similarly, based on property 3, different swapping operations will generate different $PCM^r$, which represents the transpose operation in section 2. In backward pass, the gradients respecting to $w_{S_r}$ is already obtained in equation (22). Similar to what we present in section 2, we define a backward zero matrix $\alpha_{R,R_{r-1}}^{Back}$ with a size of $(n \times \frac{(n+r-1)!}{r!(n-1)!})$. Then, we access the elements of $\alpha_{R,R_{r-1}}^{Back}$ by $PCMs^r$ as a column and plus the $w_{S_r}$ to accomplish the similar operation in (9). The operation is formulated as

$$\alpha_{R,R_{r-1}}^{Back}[PCMs^r[t]] := \alpha_{R,R_{r-1}}^{Back}[PCMs^r[t]] + w_{S_r} \quad (24)$$

where $t = 1, 2, 3, \ldots, r$. In general, for $r > 2$, the gradient with respect to $x$, will be

$$\frac{\partial y^r(x)}{\partial x} = w_{S_1} + \sum_{r=2}^{r} \alpha_{R,R_{r-1}}^{Back} x_{R_{r-1}} \quad (25)$$

Please note that the shape of $w_{S_r}$, $x_{R_{r-1}}$ and $\alpha_{R,R_{j-1}}^{Back}$ have shape $(\frac{(n+r-1)!}{r!(n-1)!} \times 1)$, $(\frac{(n+r-1)!}{r!(n-1)!} \times 1)$ and $(n \times \frac{(n+r-1)!}{r!(n-1)!})$ in this section, which is different from section 2.

### 3.3 High-order convolution implementation

We use Pytorch framework to implement our algorithms. Before creation of higher order terms, we use img2col function to convert data patches into columns, then simply follow the process we mentioned in (22) to obtain higher order terms progressively. After that, the higher order convolution is working as convolution which only needs to define the higher order kernel size with stride 1. In backpropagation, the gradient of $\frac{\partial E}{\partial w}$ and $\frac{\partial E}{\partial x_{S_r}}$ will be computed automatically since the higher order terms are inputs of conventional layer. According to the chain rule, the gradient with respect to input is $\frac{\partial E}{\partial x} = \frac{\partial E}{\partial x_{S_r}} \frac{\partial x_{S_r}}{\partial x}$ can be obtained by similar operations of (25) and (26). We also define a backward zero matrix $\alpha_{R,R_{r-1}}^{Back}$ with a size of $(n \times \frac{(n+r-1)!}{r!(n-1)!})$. Then, we access the elements of $\alpha_{R,R_{r-1}}^{Back}$ by $PCMs^r$ as a column and plus the $\frac{\partial E}{\partial x_{S_r}}$. Finally, $\frac{\partial E}{\partial x}$ is

$$\frac{\partial E}{\partial x} = \sum_{r=2}^{r} \alpha_{R,R_{r-1}}^{Back} x_{S_{r-1}} \quad (26)$$

Finally, an appropriate inverse col2im function will used to obtain input gradients $\frac{\partial E}{\partial x}$.

## 4. Complexity analysis and limitations

In this section, the time complexity of algorithm 1 and algorithm 2 will not be covered since those processes are pre-processes. Besides, $PM_{full}^r$ can be deleted if we have obtained $PCMs^r$. The theoretical comparison of time and space complexity between Traditional Volterra-based Convolution (TVC) and our Efficient Volterra-based Convolution (EVC) is demonstrated in the Table 1 (note that time complexity of accessing to elements has not included in followed table). Figures 1 and 2 illustrate the theoretical and actual speed, as well as space consumption comparison, on CPU.

|  | Time complexity | Space complexity |
|---|---|---|
| TVC forward pass | $\mathcal{O}\left(\sum_{r=1}^{r} n^r\right)$ | $\mathcal{O}\left(\sum_{r=1}^{r} n^r\right)$ |
| EVC forward pass | $\mathcal{O}\left(\frac{(n+r)!}{n!\,r!}\right)$ | $\mathcal{O}\left(\frac{(n+r)!}{n!\,r!}\right)$ |
| TVC backward pass | $\mathcal{O}\left(\sum_{r=1}^{r} n^r\right)$ | $\mathcal{O}\left(\sum_{r=1}^{r} n^r\right)$ |
| EVC backward pass | $\mathcal{O}\left(\frac{(n+r)!}{n!\,r!}\right)$ | $\mathcal{O}\left(\frac{r(n+r)!}{n!\,r!}\right)$ |

Table 1. Theoretical Time and space complexity comparison

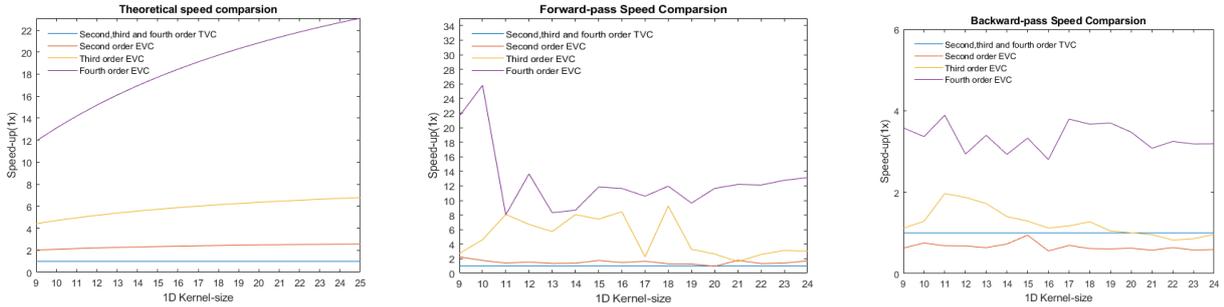

Fig 1. The theoretical and actual speed comparison for second, third and fourth order Volterra convolution between EVC and TVC. Second, third and fourth order TVC as base line (blue lines in the figure). The batch, number of input channels and number of output channel are 10.

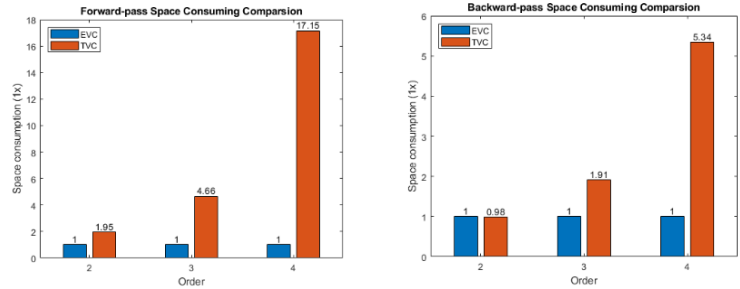

Fig 2. Space consumption comparison. The number of input channels is scaled up to 100 and the kernel size is 25.

Our EVC has higher Space complexity because we need $PCMs^r$ to accomplish forward and backward pass. Therefore, our EVC will lead to a limitation that EVC is only suitable for small convolutional kernel. However, the space complexity of EVC will become insignificant in CNNs because CNN models process multidimensional data, and small kernel sizes are widely used. For example, consider multi-dimension input is $I \in R^{b \times c \times n}$, the space complexity of $PCMs^r$ is $\mathcal{O}\left(r\frac{n(n+r)!}{n!\,r!}\right) \ll \mathcal{O}\left(bc\frac{n(n+r)!}{n!\,r!}\right)$, if $b$ and $c$ are much larger than $r$. Although forward pass of EVC is slower than theoretical comparison, the speed is also faster than TVC. In backward pass, our second order EVC is slower than second order TVC because the weights will be accessed in EVC and TVC only needs to transpose the weights. The reason that third and fourth order backward pass of EVC is faster than TVC is the number of multiplications is increased (much less than TVC) that covered the time consuming of accessing to elements. In general, our EVC acceleration depends on the time accessing to high-order terms and weights. The space consumption almost satisfies the theoretical tendency that the higher order saves the space more.

## 5. High-order Local Attention Block (HLA)

Fig.3 shows the diagram of HLA block that we propose in this study. This block is inspired by the Squeeze-And-Excitaion(SE) block, shake-shake regularization and shake drop regularization [19, 28, 29]. Basically, the SE block is adding an attention coefficient for each channel. However, the attention coefficient may fall into minimum value since the gradient of sigmoid function nearly vanishes when input is a large negative value. This condition may lead to losing some feature maps that is a problem for early convolution layers with relatively small number of channels. Besides, SE only performs channel attention which lacks the spatial attention that reduces effectiveness of attention. One of widely used spatial attention method is fusing the different attention feature maps [30–32]. Although those spatial attentions obtain impressive effectiveness, the interaction of region in each feature maps are often ignored. Therefore, we employ the Volterra convolution to provide the (nonlinear) interaction information for local attention (as shown in right branch of Fig. 3). As mentioned earlier, the right branch of HLA has same problem that may lose local information since the use of sigmoid functions. To solve this problem, we employ the idea of shake-shake regularization and shake drop regularization that use two branches to enhance local attention coeffients. In shake-shake regularization, two convlutional branches are used to reduce the overfitting. Shake drop regularization proposed a simple calculation process $y = a + b - ab, a \epsilon (0,1), b \epsilon (0,1)$ that has special functionality $y \epsilon (\max(a,b), 1)$. Based on characteristic shake-shake regularization and shake drop regularization, we use SE as global attention related to local attention by calculation process of shake drop. In the SE branch, we keep the value below channel mean and change the value over channel mean to channel mean. This process is used to prevent local attention coefficient nearly to 1 because $y = a + b - ab$ will increase the value of $\max(a,b)$.

In Fig .3, the CM is the channel mean, $r$ is the channel reduction ratio, and the Volterra Conv is second order convolution with kernel size 3×3, stride 1 and padding 1×1. The Batch Norm in the right branch can be removed if the input is the output of Batch Norm. Sigmoid block is Sigmoid function.

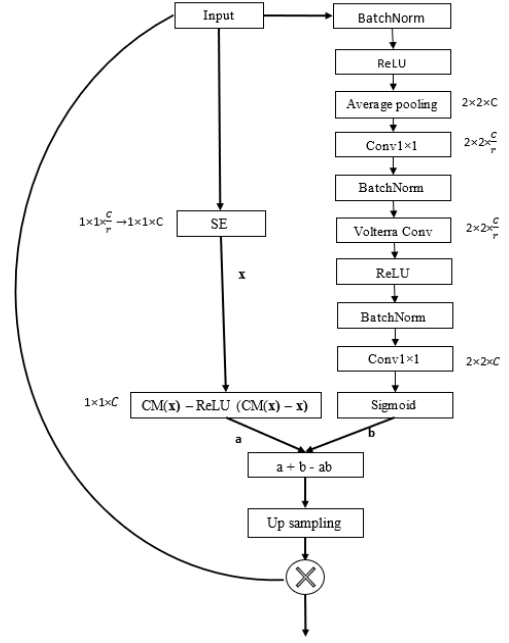

Fig.3 HLA block

## 6. Experiments and Results

In this section, we investigate the effectiveness of our HLA block applying in Wide Residual Networks (WRNs) and Pyramid Networks [28-30], on dataset Cifar-100[27]. For all WRNs, the down sampling was performed by the second layers of convolution groups and no drop-out was used. The learning rate, weight decay and the number of training epoch were the same as original WRNs. In the first and second stage of WRN-16-8, each group was followed an HLA block. In the last stage, each group was followed a normal SE block. In the experiment of WRN28-10, we found that adding more groups (with SE block) in last stage would reduce the accuracy. Therefore, we modified original WRN28-10 which have 6 groups in second stage and 2 groups in last stage. The strategy of adding HLA block was the same as WRN-16-8. The channel reduction ratio of HLA was 16 for all WRNs.

In PyramidNet-110 $\alpha$270, we found that in both first and second stages, every 4 groups are followed by an HLA block. This configuration yielded the best result, as shown in PyramidNet-110 $\alpha$270 + Shakedrop + HLA. Based on observations of WRN28-10, adding more convolution layers in the last stage did not help the network. Therefore, we modified the original PyramidNet-110 $\alpha$270 that used two basic Residual Blocks instead of last stage of PyramidNet-110 $\alpha$270. Besides, we increased the number of Pyramid blocks to 22 in both first and second stage with

$\alpha = 330$. Then, HLA-PRE (moved before the residual unit) was used when input channels were over 32. The channel reduction ratio was 8 for all PyramidNet. All training parameters and training epochs were the same as original PyramidNet report (Modified PyramidNet + Shakedrop + AA + cutout + HLA extends 30 epochs with learning rate 0.0005). Experiment results are demonstrated in Table 1.

| Cifar-100 | | | | |
|---|---|---|---|---|
| **Networks** | **Parameters** | **Augmentation** | **Attention** | **Top-1** |
| WRN16-8 | 11.0M | Standard | NO | 78.92 |
| WRN16-8 | - | Standard | SE | 80.86 |
| WRN16-8 | 11.15M | Standard | HLA + SE | 81.5±0.3 |
| WRN16-8 | 11.15M | AA + cutout | HLA + SE | 83.3±0.2 |
| WRN28-10 | 36.5M | Standard | NO | 80.75 |
| Volterra-based WRN28-10 [14] | 36.7M | Standard | NO | 81.76 |
| Modified WRN28-10 | 26.3M | Standard | HLA + SE | **82.7±0.2** |
| WRN28-10 | 36.5M | AA + cutout | NO | 82.9±0.3 |
| Modified WRN28-10 | 26.3M | AA + cutout | HLA + SE | **84.59** |
| PyramidNet-110 $\alpha$270 + Shakedrop | 28.3M | Standard | NO | 83.78 |
| PyramidNet-110 $\alpha$270 + Shakedrop | 28.4M | Standard | HLA | **84.4** |
| Modified PyramidNet + Shakedrop | 24.0M | Standard | HLA | **84.84** |
| Modified PyramidNet- + Shakedrop | 24.0M | AA + cutout | HLA | **87.12** |

Table 1. Experiment results on Cifar-100. AA is auto-augmentation [33], cutout is cutout augmentation [34] and standard is standard augmentation in WRNs. Besides, shake drop is in batch-level.

## 7. Conclusion

In this paper, we have proposed a novel and efficient method to perform higher-order Volterra filtering in DCNNs. With the order increased, our method will dramatically reduce the memory requirement that may make higher-order convolution possibly apply in CNNs or other models. Although our method for backward pass of second order convolution has slower computation speed, the higher order convolutions ($r > 2$) perform computation speed advantage. Our method can be optimized by accelerating operation that accessing the elements of matrix. Besides, we propose HLA attention module presenting that higher-order convolution can also be used in attention modules. This has been demonstrated in the classification tasks of Cifar-100 dataset. In the future, we will test out methods in different datasets and different tasks including medical image segmentation and classification for disease diagnosis and prognosis. We will also further investigate how to effectively use higher order convolution in DCNNs' architecture.